\documentclass{article}

\usepackage{arxiv}

\usepackage[utf8]{inputenc}
\usepackage[T1]{fontenc}
\usepackage{hyperref}
\usepackage{url}
\usepackage{booktabs}
\usepackage{amsfonts}
\usepackage{nicefrac}
\usepackage{microtype}
\usepackage{lipsum}
\usepackage{graphicx}

\usepackage{amsmath,amssymb,amsfonts}

\usepackage{algorithm}
\usepackage{algorithmic}
\usepackage{textcomp}
\usepackage{xcolor}
\usepackage{pifont}
\usepackage{multirow}
\usepackage[table]{xcolor}  
\usepackage{cite}

\newcommand{\xmark}{\ding{55}}

\definecolor{mygray}{gray}{0.9}

\hypersetup{
    pdfborder={0 0 0},
    colorlinks=true,
    linkcolor=black,
    citecolor=blue,
    urlcolor=blue
}

\title{EndoCIL: A Class-Incremental Learning Framework for Endoscopic Image Classification}

\author{
 Bingrong Liu \\
  School of Software\\
  Hefei University of Technology\\
  Hefei 230601, China \\
  \texttt{bingrong-liu@mail.hfut.edu.cn} \\
   \And
 Jun Shi\textsuperscript{*} \\
  School of Software\\
  Hefei University of Technology\\
  Hefei 230601, China \\
  \texttt{juns@hfut.edu.cn} \\
  \And
 Yushan Zheng\textsuperscript{*} \\
  School of Engineering Medicine\\
  Beijing Advanced Innovation Center on Biomedical Engineering\\
  Beihang University\\
  Beijing 100191, China \\
  \texttt{yszheng@buaa.edu.cn} \\
}

\renewcommand{\shorttitle}{\textit{arXiv} Template}

\begin{document}
\maketitle
\begin{abstract}
Class-incremental learning (CIL) for endoscopic image analysis is crucial for real-world clinical applications, where diagnostic models should continuously adapt to evolving clinical data while retaining performance on previously learned ones.
However, existing replay-based CIL methods fail to effectively mitigate catastrophic forgetting due to severe domain discrepancies and class imbalance inherent in endoscopic imaging.
To tackle these challenges, we propose EndoCIL, a novel and unified CIL framework specifically tailored for endoscopic image diagnosis.
EndoCIL incorporates three key components: Maximum Mean Discrepancy Based Replay (MDBR), employing a distribution-aligned greedy strategy to select diverse and representative exemplars, Prior Regularized Class Balanced Loss (PRCBL), designed to alleviate both inter-phase and intra-phase class imbalance by integrating prior class distributions and balance weights into the loss function, and Calibration of Fully-Connected Gradients (CFG), which adjusts the classifier gradients to mitigate bias toward new classes.
Extensive experiments conducted on four public endoscopic datasets demonstrate that EndoCIL generally outperforms state-of-the-art CIL methods across varying buffer sizes and evaluation metrics.
The proposed framework effectively balances stability and plasticity in lifelong endoscopic diagnosis, showing promising potential for clinical scalability and deployment.
\end{abstract}


\section{Introduction}
\begin{figure}[ht]
    \centering
    \includegraphics[width=0.9\textwidth]{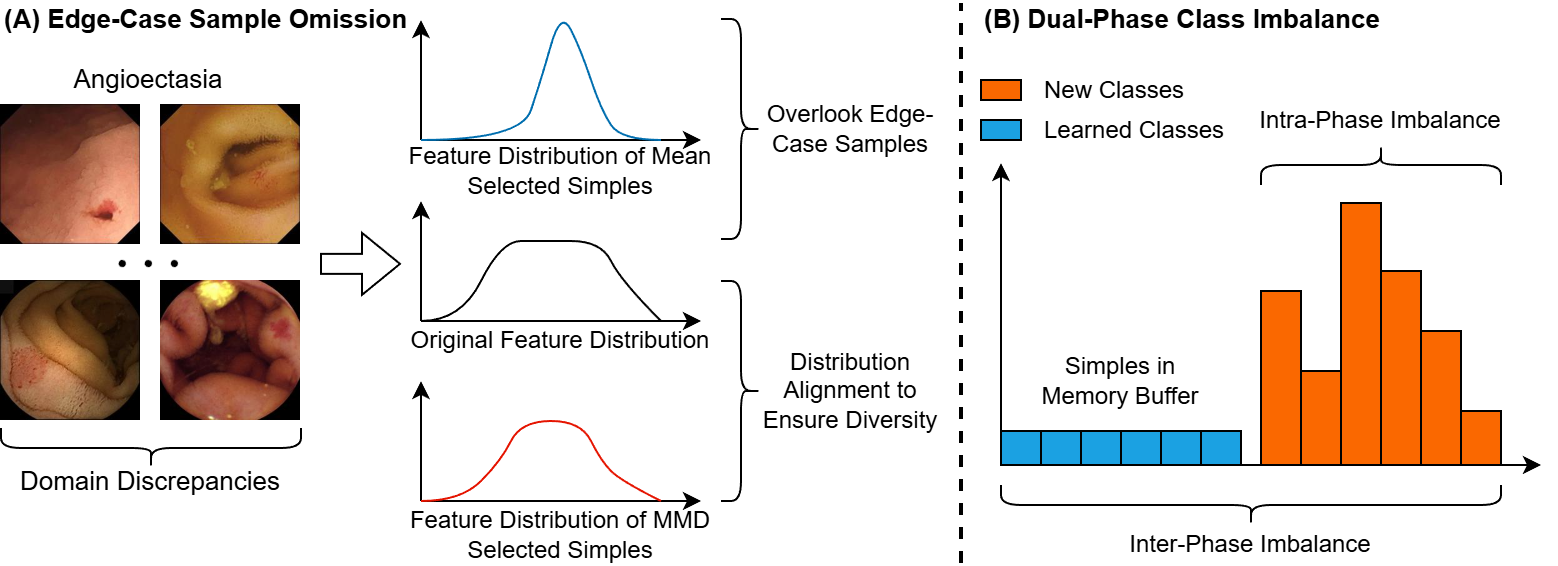}
    \caption{Illustration of key challenges: (A) shows the limitations of the mean exemplar selection strategy, addressed via MMD-Based Replay to preserve feature diversity; (B) highlights the dual-phase class imbalance, mitigated by Prior Regularized Class Balanced Loss and Calibration of Fully-Connected Gradients.}
    \label{img:issue}
\end{figure}

Gastrointestinal cancer is one of the leading causes of morbidity and mortality worldwide.
Early screening and timely treatment are crucial for reducing both its incidence and fatality rates~\cite{banik2020polyp, yang2020colon}.
With the rapid advancement of artificial intelligence, computer-aided diagnosis (CAD) systems have become indispensable tools in endoscopic examinations, assisting physicians in detecting lesions and diagnosing diseases with higher efficiency and accuracy~\cite{caroppo2021deep, ozawa2019novel}.
However, most of existing CAD models are trained on static datasets and assume to be executed on a fixed data distribution.
This approach fails to address the evolving nature of clinical data, where new disease types, imaging modalities, and diagnostic knowledge are continuously emerging.
As a result, these models are inherently limited in real-world applications where continual updates are essential~\cite{wu2024continual, zhang2023adapter}.

In endoscopy-based diagnosis, continually adapting models to newly collected data is especially important.
New disease categories, variants of existing ones, and different anatomical landmarks may emerge, requiring diagnostic models to expand their recognition capabilities.
Rare cases, which may be excluded during initial training due to insufficient samples, should also be introduced to the model as their sample size increases over time.
Moreover, endoscopic images acquired from different devices often exhibit significant domain shifts due to variations in camera optics, imaging principles, and acquisition parameters.
These differences can intuitively manifest in images' statistical characteristics, such as RGB or HSV color space, brightness, contrast, and resolution~\cite{xu2022deep}, posing a challenge for conventional models trained in closed-world settings.
Therefore, there is an urgent need to develop more advanced and flexible learning frameworks that can adapt to the dynamic nature of endoscopic data.

Recently, continual learning, particularly class-incremental learning (CIL), has gained increasing attention in the medical imaging domain~\cite{ayromlou2024ccsi, zhu2024lifelong}.
It aims to sequentially incorporate new class knowledge while retaining performance on previously learned categories, thereby achieving a balance between model plasticity and stability.
Among existing CIL approaches, replay-based methods have shown promising results by storing a small number of exemplars from previous tasks for experience replay~\cite{rebuffi2017icarl, chaudhry2018riemannian}.
However, current replay-based methods are designed for natural image domains and fail to consider the specific characteristics of endoscopic data.
For instance, they often rely on exemplar selection strategies based on feature means or prediction entropy.
Yet, due to the intrinsic domain discrepancies of endoscopic imagery, such metrics may not guarantee the diversity and representativeness of the replay buffer.
These selection strategies typically optimize a single evaluation criterion and ignore the overall feature distribution, potentially discarding important edge-case samples.
Moreover, many existing methods~\cite{wang2022foster, zhou2022model} assume a balanced data distribution.
In contrast, endoscopic datasets are inherently imbalanced due to clinical reality—only a small portion of the gastrointestinal tract typically contains lesions~\cite{yue2023automated}.
In the context of CIL, this imbalance manifests in two distinct forms:
\textbf{intra-phase imbalance}, where the distribution of current classes is skewed, and 
\textbf{inter-phase imbalance}, caused by discrepancies between newly introduced samples and exemplars retained from earlier phases~\cite{he2024gradient}.
The former often causes the model to overfit the majority classes while underrepresenting minority and previously learned ones, and the latter biases the classifier toward newly introduced classes~\cite{chen2023dynamic}.
This dual imbalance severely degrades the performance of CIL models and exacerbates catastrophic forgetting.
These challenges are illustrated in Fig.~\ref{img:issue}, where (A) shows limitations of exemplar selection strategies and (B) highlights the dual-phase imbalance.

To overcome these limitations, we propose a novel endoscopic class-incremental learning framework, named EndoCIL.
The main contributions of this work are as follows:
\begin{itemize}
\item We propose a novel class-incremental learning framework specifically designed for endoscopic image analysis (EndoCIL). To the best of our knowledge, this is the first work to tackle class-incremental learning in the domain of endoscopic imagery.
\item We propose a unified strategy to address exemplar selection and class imbalance in incremental learning. Specifically, we introduce Maximum Mean Discrepancy~\cite{gretton2012kernel} Based Replay (MDBR) to select exemplars that preserve the original feature distribution. To tackle inter-phase and intra-phase imbalance, we design Prior Regularized Class Balanced Loss (PRCBL), which incorporates prior statistics and class balanced weights. We further develop Calibration of Fully-Connected Gradients (CFG) to reduce classifier bias by modulating gradients in the fully connected layer.
\item Experiments on four publicly available endoscopic image datasets demonstrate that our method outperforms existing state-of-the-art class-incremental learning approaches in endoscopic image class-incremental learning tasks.
\end{itemize}

\section{Related Work}
\subsection{Class-Incremental Learning} 
Class-incremental learning (CIL) is a subfield of continual learning in which a model is required to incrementally acquire new classes over time, while retaining the ability to recognize previously learned categories.
Crucially, during inference, task identity is not provided, and the model must predict over all seen classes using a unified single-head classifier~\cite{maltoni2019continuous}. 

Conventional CIL approaches can be broadly categorized into three main families: regularization-based methods, replay-based methods, and parameter-isolation-based methods.
Regularization-based methods aim to mitigate forgetting by restricting updates to parameters deemed important for previous tasks.
For instance, EWC~\cite{kirkpatrick2017overcoming} estimates the Fisher information matrix to quantify parameter importance, thereby penalizing updates that might interfere with previously acquired knowledge.
Other representative techniques rely on knowledge distillation from output logits~\cite{li2017learning} or intermediate feature representations~\cite{douillard2020podnet} to retain performance on old classes.
Replay-based methods maintain a small memory buffer containing representative exemplars from past tasks.
These samples are reused during training of new tasks to preserve prior knowledge.
iCaRL~\cite{rebuffi2017icarl} uses class mean-based selection to construct the memory buffer.
Later, Chaudhry et al.~\cite{chaudhry2018riemannian} proposed a selection strategy based on prediction entropy and proximity to the decision boundary, encouraging the inclusion of more informative and uncertain samples.
Parameter-isolation methods tackle forgetting by explicitly reserving separate network subspaces or expanding model capacity for each task.
Techniques such as dynamically expandable networks~\cite{yan2021dynamically} aim to protect prior task parameters.
However, these methods often incur high computational and memory overhead, limiting their scalability in long-term deployment scenarios.

Compared to standard CIL settings, imbalanced CIL has received less attention due to its dual challenges: mitigating catastrophic forgetting while addressing class imbalance across and within incremental phases.
Initial efforts addressed imbalanced scenarios in multi-label~\cite{kim2020imbalanced} or semi-supervised learning settings~\cite{belouadah2020active}.
More recent method~\cite{liu2022long} adopts a decoupled two-stage strategy to separate representation learning from classifier adaptation in long-tailed incremental setups.
Chen et al.~\cite{chen2023dynamic} later propose dynamic classifier calibration to address distribution shifts between old and new classes.
He et al.~\cite{he2024gradient} further tackle this issue by reweighting the gradients of the classifier and introducing a distribution-aware knowledge distillation loss.

\subsection{Maximum Mean Discrepancy}

Maximum Mean Discrepancy (MMD)~\cite{gretton2012kernel} is a non-parametric statistical metric used to evaluate the distance between two probability distributions in a reproducing kernel Hilbert space (RKHS).
Given a positive definite kernel $k: \mathcal{X} \times \mathcal{X} \rightarrow \mathbb{R}$, there exists a feature mapping $\phi: \mathcal{X} \rightarrow \mathcal{H}$ into an RKHS $\mathcal{H}$ such that $k(x, y) = \langle \phi(x), \phi(y) \rangle_{\mathcal{H}}$.
The kernel mean embedding of a probability distribution $P$ is defined as:
\begin{equation}
\mu_P = \mathbb{E}_{x \sim P}[\phi(x)]
\end{equation}

The MMD between two distributions $P$ and $Q$ is then given by the distance between their kernel mean embeddings:
\begin{equation}
\mathrm{MMD}(P, Q) = \left\| \mu_P - \mu_Q \right\|_{\mathcal{H}}
\label{eq:h-mmd}
\end{equation}

If the kernel $k$ is characteristic, this mapping is injective and the MMD defines a proper metric on the space of distributions.
In practical applications, MMD is estimated empirically from finite independent and identically distributed samples. For instance, given two sample sets $\mathcal{X} = \{x_i\}_{i=1}^{m}$ from distribution $P$ and $\mathcal{Y} = \{y_j\}_{j=1}^{n}$ from $Q$, the empirical MMD is computed as:

\begin{equation}
\mathrm{MMD}^2(\mathcal{X}, \mathcal{Y}) = \frac{1}{m^2} \sum_{i=1}^{m} \sum_{j=1}^{m} k(x_i, x_j) + \frac{1}{n^2} \sum_{i=1}^{n} \sum_{j=1}^{n} k(y_i, y_j) - \frac{2}{mn} \sum_{i=1}^{m} \sum_{j=1}^{n} k(x_i, y_j)
\end{equation}

MMD has been widely used in domain adaptation, generative modeling, and privacy-preserving data analysis.
In the context of CIL, MMD provides a principled way to assess distributional alignment between the stored exemplar set and the original data.

\section{Methodology}
\begin{figure*}[ht] 
    \centering
    \includegraphics[width=0.9\textwidth]{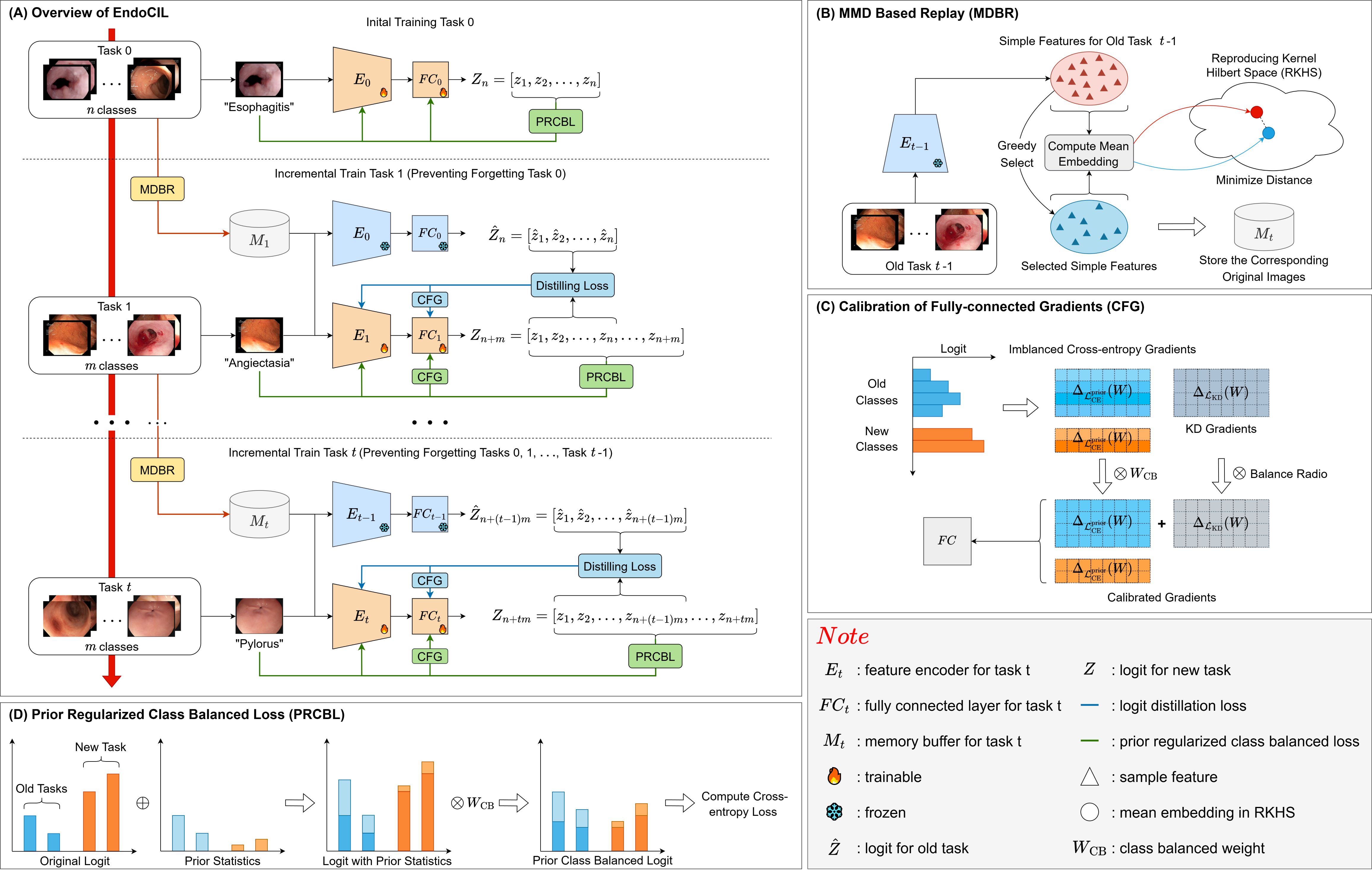}
    \caption{
    Overview of the proposed method. When task $t = 0$, the model is trained from scratch. For each subsequent task ($t > 0$), a representative subset is selected via the MDBR (B) to approximate the feature distribution of previous tasks. Then, the data from the current task and the stored exemplars in the memory buffer are jointly used as training input. During training, the logits from the previous model are aligned with those of the current model using a distillation loss. The prediction loss is computed using the PRCBL (D). The gradients induced by both losses on the fully connected (FC) layer are modulated by the CFG (C), mitigating the classifier's bias toward new classes.
    }
    \label{img:mothed}
\end{figure*}

\subsection{Overview of EndoCIL} 
Class-incremental learning (CIL) aims to progressively train a model on a sequence of tasks, where each task introduces a disjoint set of new classes.
Formally, the training data for the $t$-th task is denoted as $\mathcal{D}^t = \{(x_{i,t}, y_{i,t})\}_{i=1}^{m_t}$, where $t \in \{0, 1, \dots, T\}$ indexes the current task and $m_t$ is the number of samples in task $t$.
The objective is to enable the model to recognize all classes encountered up to the current task, such that the cumulative label space becomes $\mathcal{Y}_1 \cup \cdots \cup \mathcal{Y}_t$.

As illustrated in Fig.~\ref{img:mothed}, the model is initially trained from scratch on task $t=0$.
For each subsequent task ($t > 0$), a representative subset of previous data is selected via our MDBR (B) to approximate the feature distribution of previous tasks.
The model is then updated using both the current task's data and stored exemplars from the memory buffer.
To mitigate forgetting, we apply knowledge distillation by aligning the logits of the current model with those of the previous one.
Meanwhile, the prediction loss is computed using our proposed PRCBL (D), which enhances class-balanced learning.
Finally, the gradients induced by both losses on the fully connected (FC) layer are modulated by the CFG (C), thereby mitigating the classifier's bias toward new classes.

\subsection{Maximum Mean Discrepancy Based Replay}

\subsubsection{Problem Definition}

Let $\mathcal{X}_c = \{x_1, x_2, \dots, x_m\} \subset \mathbb{R}^d$ denote the set of normalized feature vectors extracted from all training samples belonging to class $c$, where $m$ is the total number of samples and $d$ is the feature dimension.
We aim to select a subset $\mathcal{S}_c \subset \mathcal{X}_c$, with $|\mathcal{S}_c| = n < m$, such that the squared MMD between $\mathcal{S}_c$ and $\mathcal{X}_c$ is minimized:

\begin{equation}
\mathrm{MMD}^2(\mathcal{X}_c, \mathcal{S}_c) = \frac{1}{m^2} \sum_{x, x' \in \mathcal{X}_c} k(x, x') +  \frac{1}{n^2} \sum_{s, s' \in \mathcal{S}_c} k(s, s')\\ - \frac{2}{mn} \sum_{x \in \mathcal{X}_c} \sum_{s \in \mathcal{S}_c} k(x, s)
\label{eq:mmd}
\end{equation}

Here, $k(\cdot, \cdot)$ is constructed as a sum of $K$ Gaussian Radial Basis Function (RBF) kernels with varying bandwidths:

\begin{equation}
k(x, x') = \sum_{i=1}^{K} \exp\left(-\frac{|x - x'|^2}{\sigma_i}\right)
\end{equation}
where each $\sigma_i$ is computed as $\sigma_i = \frac{d}{\mu^{K/2 - i}}$, with $d$ being the average pairwise squared distance among all samples, and $\mu$ is a multiplicative scaling factor.

Note that the first term in Eq.~\eqref{eq:mmd}, $\frac{1}{m^2} \sum_{x, x' \in \mathcal{X}_c} k(x, x')$ is constant with respect to the selection and can be precomputed. Therefore, the objective function to be minimized is reformulated as:

\begin{equation}
O(\mathcal{X}_c, \mathcal{S}_c) = \frac{1}{n^2} \sum_{s, s' \in \mathcal{S}_c} k(s, s') - \frac{2}{mn} \sum_{x \in \mathcal{X}_c} \sum_{s \in \mathcal{S}_c} k(x, s)
\label{eq:obj}
\end{equation}

\subsubsection{Greedy Selection Strategy} 

In theory, identifying the optimal subset $\mathcal{S}_c$ requires evaluating all possible combinations from $\mathcal{X}_c$, which is computationally infeasible due to the combinatorial explosion. Instead, we adopt a greedy selection strategy to efficiently approximate the optimal subset that minimizes MMD.

We begin by computing the complete kernel matrix $K \in \mathbb{R}^{m \times m}$ over all pairwise combinations in $\mathcal{X}_c$, where $K_{ij} = k(x_i, x_j)$.
This precomputation allows all required kernel evaluations during the selection process to be directly retrieved from $K$, significantly reducing computational overhead.
Then, we iteratively construct $\mathcal{S}_c$ by selecting one sample at a time over $n$ steps. 

At iteration $t$, we denote the current selected subset as $\mathcal{S}_c^{t-1}$ with $|\mathcal{S}_c^{t-1}| = t-1$, and evaluate each candidate sample $x_j \in \mathcal{X}_c \setminus \mathcal{S}_c^{t-1}$ using the following objective function:

\begin{equation}
\mathcal{F}_t(x_j) = \frac{
\sum\limits_{s,s' \in \mathcal{S}_c^{t-1}} k(s, s') 
+ 2 \cdot \sum\limits_{s \in \mathcal{S}_c^{t-1}} k(s, x_j) 
+ k(x_j, x_j)}{t^2} 
- \frac{2}{mt}  \left( \sum\limits_{x \in \mathcal{X}_c} \sum\limits_{s \in \mathcal{S}_c^{t-1}} k(x, s) 
+ \sum\limits_{x \in \mathcal{X}_c} k(x, x_j) \right)
\end{equation}
where:
\begin{itemize}
  \item $\sum_{s,s' \in \mathcal{S}_c^{t-1}} k(s, s')$ is the accumulated intra-subset kernel sum,
  \item $\sum_{s \in \mathcal{S}_c^{t-1}} k(s, x_j)$ is the kernel sum between the candidate $x_j$ and the current subset,
  \item $\sum_{x \in \mathcal{X}_c} \sum_{s \in \mathcal{S}_c^{t-1}} k(x, s)$ is the cross-kernel sum between the current subset and full data,
  \item $\sum_{x \in \mathcal{X}_c} k(x, x_j)$ is the cross-kernel sum for the candidate $x_j$.
\end{itemize}

The sample with the lowest $\mathcal{F}_t(x_j)$ is selected and added to the subset: $\mathcal{S}_c^{t} = \mathcal{S}_c^{t-1} \cup \{x_j\}$. 
This procedure is repeated until the subset reaches the desired size.
Notably, $\mathcal{F}_t(x_j)$ is derived from Eq.~\eqref{eq:obj} and represents the estimated objective value if $x_j$ were included in the current subset at iteration $t$. 
Therefore, greedily minimizing $\mathcal{F}_t(x_j)$ at each step provides an efficient approximation to the global minimization of $O(\mathcal{X}_c, \mathcal{S}_c)$. The detailed algorithm is shown in Algorithm~\ref{alg:greedy_mmd}.

\begin{algorithm}[ht]
\caption{Greedy Selection for Minimizing MMD}
\label{alg:greedy_mmd}
\begin{algorithmic}
\STATE \textbf{Input:} Feature set $\mathcal{X}_c = \{x_1, \dots, x_m\} \subset \mathbb{R}^d$, target size $n$, kernel $k(\cdot, \cdot)$
\STATE \textbf{Output:} Selected subset $\mathcal{S}_c \subset \mathcal{X}_c$, $|\mathcal{S}_c| = n$
\STATE Compute kernel matrix $K \in \mathbb{R}^{m \times m}$, $K_{ij} = k(x_i, x_j)$
\STATE Compute $s_j = \sum_{i=1}^{m} K_{ji}$ for all $j$
\STATE Initialize $\mathcal{S}_c \gets \emptyset$, $\mathcal{C} \gets \{1, \dots, m\}$, $S_{\text{inner}} \gets 0$, $S_{\text{outer}} \gets 0$, $C_j \gets 0$
\FOR{$t = 1$ to $n$}
    \STATE $f^\ast \gets \infty$, $j^\ast \gets -1$
    \FOR{each $j \in \mathcal{C}$}
        \STATE $\Delta_{\text{inner}} \gets 2 C_j + K_{jj}$, $\Delta_{\text{outer}} \gets s_j$
        \STATE $\mathcal{F}_t(x_j) \gets \dfrac{S_{\text{inner}} + \Delta_{\text{inner}}}{t^2} - \dfrac{2}{m t}(S_{\text{outer}} + \Delta_{\text{outer}})$
        \IF{$\mathcal{F}_t(x_j) < f^\ast$}
            \STATE $f^\ast \gets \mathcal{F}_t(x_j)$, $j^\ast \gets j$
        \ENDIF
    \ENDFOR
    \STATE $\mathcal{S}_c \gets \mathcal{S}_c \cup \{x_{j^\ast}\}$, $\mathcal{C} \gets \mathcal{C} \setminus \{j^\ast\}$, $S_{\text{inner}} \gets S_{\text{inner}} + 2 C_{j^\ast} + K_{j^\ast j^\ast}$, $S_{\text{outer}} \gets S_{\text{outer}} + s_{j^\ast}$
    \FOR{each $j \in \mathcal{C}$}
        \STATE $C_j \gets C_j + K_{j^\ast j}$
    \ENDFOR
\ENDFOR
\end{algorithmic}
\end{algorithm}

\subsubsection{Theoretical Error Upper Bound}

We provide a theoretical guarantee on the performance of our greedy selection algorithm.  
It can be proved that our method is equivalent to the Greedy MMD Minimization (GM) algorithm described in Algorithm~4 of the work by Pronzato et al.~\cite{pronzato2023performance}.
Based on this equivalence, we adopt the theoretical upper bound on the approximation error established for the GM algorithm to characterize the performance of our approach.

We define the target probability measure $\mu$ as the uniform empirical measure over the finite set $\mathcal{X}_c$.
Then, following Theorem 5 in~\cite{pronzato2023performance}, the squared MMD between $\mu$ and the empirical measure $\xi_n$ supported on $\mathcal{S}_c$ can be bounded as:

\begin{equation}
\mathrm{MMD}^2(\mu, \xi_n) 
\leq M_C^2 + A_C \cdot \frac{1 + \log n}{n}
\label{eq:theory_bound}
\end{equation}
where:
\begin{itemize}
    \item $M_C^2$ denotes the optimal squared MMD achievable using any probability measure supported on $\mathcal{X}_c$
    \item $A_C = K_{\max} + \tau_{1/2}^2(\mu)$ is a constant depending on kernel boundedness
    \item $\tau_{1/2}^2(\mu) = \left[ \int_{\mathcal{X}_c} \sqrt{k(x, x)}  d\mu(x) \right]^2$ 
    simplifies to the empirical form:
    \begin{equation}
    \tau_{1/2}^2(\mu) = \left( \frac{1}{m} \sum_{x \in \mathcal{X}_c} \sqrt{k(x, x)} \right)^2
    \end{equation}
\end{itemize}

In our setting, since we use a sum of $K$ RBF kernels, and each RBF kernel is bounded by 1, we obtain:
\begin{equation}
K_{\max} = \max_{x \in \mathcal{X}_C} k(x,x) = K
\end{equation}
\begin{equation}
\tau_{1/2}^2(\mu) = \left( \frac{1}{m} \cdot m \cdot \sqrt{K} \right)^2 = K
\end{equation}

Therefore, the constant simplifies to $A_C = K + K = 2K$, and the final upper bound becomes:
\begin{equation}
\mathrm{MMD}^2(\mu, \xi_n) 
\leq M_C^2 + \frac{2K(1 + \log n)}{n}
\label{eq:theory_bound_simple}
\end{equation}

This result implies that the greedy algorithm converges to the optimal MMD approximation with a rate of $\mathcal{O}(\log n / n)$.
Such a convergence guarantee makes the method theoretically sound and practically effective for class-incremental replay settings.

\subsection{Prior Regularized Class Balanced Loss}

\subsubsection{Prior Regularized Logit}

Inspired by~\cite{menon2020long}, we incorporate a log prior vector into the logits to bias model predictions toward the observed class distribution.
Specifically, the prior probability for class $i$ is defined as:
\begin{align}
p_i^{\text{prior}} &= \frac{n_i}{\sum_{j=1}^{C} n_j}, \quad i = 1, \dots, C \label{eq:prior_prob} \\
\log \mathbf{p}^{\text{prior}} &= [\log p_1^{\text{prior}}, \dots, \log p_C^{\text{prior}}]
\end{align}
where $n_i$ is the sample count for class $i$, and $C$ is the total number of classes. The output logits $\mathbf{z} \in \mathbb{R}^C$ are regularized as:
\begin{equation}
p_i = \frac{\exp(z_i + \log p_i^{\text{prior}})}{\sum_{j=1}^C \exp(z_j + \log p_j^{\text{prior}})}
\end{equation}

\subsubsection{Class Balanced Weights}

To mitigate the impact of class imbalance, we compute a class balanced weight vector $W^{\text{CB}} \in \mathbb{R}^C$ based on the effective number of samples~\cite{cui2019class}.
Given a class with $n_i$ training samples, we defined its effective number and unnormalized class weights as:
\begin{equation}
E_i = \frac{1 - \beta^{n_i}}{1 - \beta}, \quad \alpha_i = \frac{1}{E_i} = \frac{1 - \beta}{1 - \beta^{n_i}}
\end{equation}
where $\beta \in [0, 1)$ is a hyperparameter controlling the degree of down-weighting for frequent classes.

We normalize all weights to obtain the final class balanced weight vector:
\begin{equation}
W^{\text{CB}}_i = \alpha_i \cdot \frac{C}{\sum_{j=1}^{C} \alpha_j}, \quad i = 1, \dots, C
\end{equation}

The final prior regularized class balanced loss of task $t$ is defined as:
\begin{equation}
\mathcal{L}_{\text{CE}}^{\text{prior}} = 
\begin{cases}
- \log p_y, & \text{if } t = 0 \\
- W^{\text{CB}}_y \cdot \log p_y, & \text{if } t > 0
\end{cases}
\end{equation}
where $W^{\text{CB}}_y$ is the class balanced weight of the ground-truth class $y$, and $p_y$ is the predicted probability from the prior-regularized softmax output.

To preserve previously learned knowledge, we adopt the knowledge distillation~\cite{hinton2015distilling}.
The overall training objective for incremental task $t > 0$ is given by:
\begin{equation}
\mathcal{L}_t = \mathcal{L}_{\text{CE}}^{\text{prior}} + \lambda \cdot \mathcal{L}_{\text{KD}}
\end{equation}
where $\lambda$ is a trade-off parameter balancing the classification and distillation objectives.

\subsection{Calibration of Fully-Connected Gradients}

To further address the challenge of class imbalance during optimization, we calibrate the gradients of the fully connected (FC) layer using the class balanced weight vector $W^{\text{CB}} \in \mathbb{R}^{C}$.
This mechanism ensures that each class contributes fairly to the update of classification weights, particularly under class-incremental learning scenarios.

The gradients of the FC layer parameters $\mathbf{w}_i \in \mathbb{R}^d$ and bias $b_i \in \mathbb{R}$ for each class $i \in \{1, \dots, C\}$ are computed as:
\begin{equation}
\mathbf{g}^{\text{clf}}_i = \nabla_{\mathbf{w}_i} \mathcal{L}_{\text{CE}}^{\text{prior}}, \quad 
\mathbf{g}^{\text{kd}}_i = \nabla_{\mathbf{w}_i} \mathcal{L}_{\text{KD}}
\end{equation}
\begin{equation}
b^{\text{clf}}_i = \nabla_{b_i} \mathcal{L}_{\text{CE}}^{\text{prior}}, \quad 
b^{\text{kd}}_i = \nabla_{b_i} \mathcal{L}_{\text{KD}}
\end{equation}

To balance the magnitude of classification and distillation gradients, we compute a balance ratio $\gamma$ following~\cite{he2024gradient}:
\begin{equation}
\gamma = \frac{\left\| \sum_{i=1}^C W^{\text{CB}}_i \cdot \mathbf{g}^{\text{clf}}_i  \right\|}{\left\| \sum_{i=1}^{C_{\text{old}}}  \mathbf{g}^{\text{kd}}_i \right\|}
\end{equation}

The final calibrated gradient used for updating each class $i$ in the FC layer is given by:
\begin{equation}
\tilde{\mathbf{g}}_i = W^{\text{CB}}_i \cdot \mathbf{g}^{\text{clf}}_i + \gamma \cdot \mathbf{g}^{\text{kd}}_i \cdot \mathbb{I}(i \leq C_{\text{old}})
\end{equation}
\begin{equation}
\tilde{b}_i = W^{\text{CB}}_i \cdot b^{\text{clf}}_i + \gamma \cdot b^{\text{kd}}_i \cdot \mathbb{I}(i \leq C_{\text{old}})
\end{equation}

This gradient calibration effectively preserves knowledge of previously learned classes while encouraging balanced learning of under-represented new classes.

\section{Experiments}

\subsection{Datasets}
To comprehensively evaluate the effectiveness of proposed method, we conduct experiments on four publicly available endoscopic image datasets: Hyper-Kvasir~\cite{borgli2020hyperkvasir}, Gastrovision~\cite{jha2023gastrovision}, Kvasir-Capsule~\cite{smedsrud2021kvasir} and Capsule Vision~\cite{handa2024capsule}. 

\subsubsection{Hyper-Kvasir}
Hyper-Kvasir is the largest publicly available dataset for gastrointestinal tract images.
It includes 110,079 images and 373 videos, covering a wide array of anatomical landmarks, as well as both pathological and normal findings.
The dataset is categorized into four main types: labeled image data, unlabeled image data, segmented image data, and annotated video data.
For our experiments, we focus on the labeled image data, which consists of 10,662 images across 23 distinct classes.

\subsubsection{Gastrovision}
Gastrovision is a multi-class endoscopy image dataset consisting of 8,000 images across 27 classes.
It covers two major categories—upper GI and lower GI—and includes a variety of anatomical landmarks, pathological abnormalities, polyp removal cases, and normal findings within the gastrointestinal tract.

\subsubsection{Kvasir-Capsule}
Kvasir-Capsule is a specialized dataset derived from video capsule endoscopy (VCE).
It contains 47,238 labeled images and 117 videos, with a focus on anatomical landmarks, pathological conditions, and normal findings within the gastrointestinal tract.
The dataset is divided into three categories: labeled images, labeled videos, and unlabeled videos.
For our experiments, we specifically utilize the labeled image data, which include 47,238 images across 14 distinct classes.

\subsubsection{Capsule Vision}
Capsule Vision is the dataset for the "Capsule Vision Challenge 2024: Multi-Class Abnormality Classification for Video Capsule Endoscopy."
The dataset is developed by combining portions of three publicly available VCE datasets—the SEE-AI project dataset~\cite{yokote2024small}, KID~\cite{koulaouzidis2017kid}, and Kvasir-Capsule dataset—along with one private dataset, AIIMS.
It consists of 53,739 VCE frames annotated with 10 distinct class labels.

\begin{figure*}[h]
    \centering
    \includegraphics[width=\textwidth]{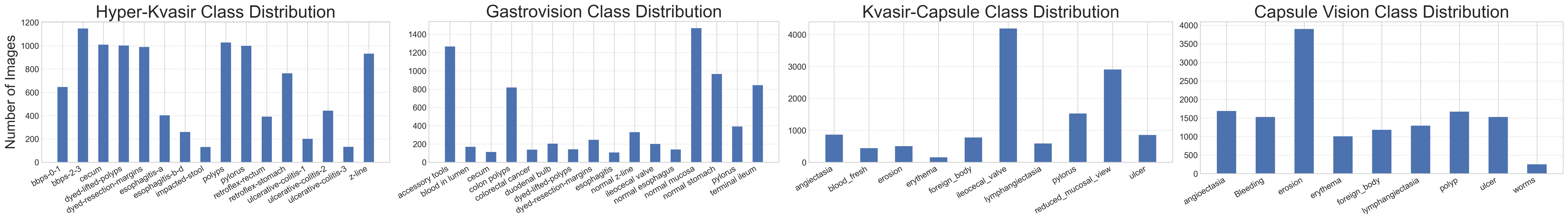}
    \caption{Class samples distribution of different datasets.}
    \label{img:class_number_distribution}
    \vspace{-1.0em}
\end{figure*}

To construct reliable class-incremental learning tasks, we exclude all classes that contain fewer than 100 images from each dataset. 
In addition, we remove the \textbf{normal clean mucosa} class from the Kvasir-Capsule dataset and the \textbf{normal} class from the Capsule Vision dataset, as their excessive sample sizes would lead to prohibitive computational overhead for kernel functions. 
The class distribution of the filtered datasets is illustrated in Fig.~\ref{img:class_number_distribution}, while retaining a certain degree of class imbalance to reflect the real-world characteristics of the datasets.

After filtering, we evenly divided the classes of each dataset into five incremental tasks, the specific split results are shown in Table.~\ref{tab:dataset_split}.
For all datasets, the training and test sets were partitioned with a ratio of 8:2.

\begin{table}[htbp]
\centering
\caption{Class division for incremental tasks across different datasets.}
\begin{tabular}{lccc}
\hline
\textbf{Dataset}       & \textbf{Classes} & \textbf{Tasks} & \textbf{Classes per task} \\ \hline
Hyper-Kvasir   & 16       & 5     & {[4, 3, 3, 3, 3]} \\
Gastrovision     & 16       & 5     & {[4, 3, 3, 3, 3]} \\
Kvasir-Capsule & 10       & 5     & {[2, 2, 2, 2, 2]} \\
Capsule Vision & 9       & 4     & {[3, 2, 2, 2]} \\ \hline
\end{tabular}
\label{tab:dataset_split}
\end{table}

\subsection{Experimental Setting} 
All experiments are implemented using the PyCIL framework~\cite{zhou2023pycil} with the same hyperparameters.
We follow the official configuration settings along with several recent works based on this platform~\cite{bian2024make, zhou2024class}.
Specifically, ResNet18~\cite{he2016deep} is adopted as the backbone feature extractor due to its wide adoption in most existing replay-based CIL methods.
All input images are resized to $256 \times 256$ before being fed into the network.
The initial model on the first task is trained from scratch using stochastic gradient descent (SGD) for 100 epochs, with a learning rate of 0.1, momentum of 0.9, and a weight decay of $5 \times 10^{-4}$.
The learning rate decays by a factor of 0.1 at epochs 50 and 70, and the batch size is fixed to 64.
In the subsequent incremental phases, the training setup remains largely the same, except the weight decay is reduced to $2 \times 10^{-4}$.
Moreover, the distillation temperature $T$ is set to 2 for methods incorporating knowledge distillation during the incremental phases.
For the hyper-parameters of our method, we set the effective number ratio $\beta$ to 0.96 and the distillation loss balancing coefficient $\lambda$ to 1.
All experiments were conducted using PyTorch 2.1.0 with CUDA 12.4 on Ubuntu 22.04, running on a workstation equipped with a single NVIDIA RTX 3090 GPU and 90 GB of RAM.

Following~\cite{zhou2023pycil, zhou2024class, zheng2024multi}, we report three standard metrics commonly used in class-incremental learning: the accuracy of the last task ($Acc_{\text{last}}$), the average accuracy across all tasks ($Acc_{\text{avg}}$), and the average forgetting ($AF$), which measures the performance drop on previously learned tasks.
Specifically, we record the accuracy of each task immediately after training and after the final task, and $AF$ is computed as the average difference across all tasks. 
$Acc_{\text{avg}}$ and $AF$ are formulated as follows: 
\begin{equation}
Acc_{\text{avg}} = \frac{1}{T + 1} \sum_{t=0}^{T} Acc(t)
\end{equation}
\begin{equation}
AF = \frac{1}{T + 1} \sum_{t=0}^{T} (Acc(t) - Acc_{\text{last}}(t))
\end{equation}
where $Acc(t)$ represents the accuracy after completing the $t$-th task, and $Acc_{\text{last}}(t)$ is the accuracy of the $t$-th task after the final task.

In addition to these metrics, we also report commonly used indicators for endoscopic image classification, including the F1-score of the last task ($F1_{\text{last}}$) and the average F1-score across all tasks ($F1_{\text{avg}}$). 
$F1_{\text{avg}}$ is formulated as follows:
\begin{equation}
F1_{\text{avg}} = \frac{1}{T + 1} \sum_{t=0}^{T} F1(t)
\vspace{-1.0em}
\end{equation}

\subsection{Quantitative Results}

\begin{table*}[ht]
\centering
\caption{Comparison of class-incremental learning methods on four endoscopic datasets. For replay-based methods under different buffer sizes, the best results are shown in \textbf{bold}, and the second-best are \underline{underlined}.}
\resizebox{\textwidth}{!}{
\begin{tabular}{l|ccccc|ccccc|ccccc|ccccc}
\toprule
\textbf{Method} & \multicolumn{5}{c|}{\textbf{Hyper-Kvasir}} & \multicolumn{5}{c|}{\textbf{Gastrovision}} & \multicolumn{5}{c|}{\textbf{Kvasir-Capsule}} & \multicolumn{5}{c}{\textbf{Capsule Vision}} \\
 & $Acc_{\text{last}}$ & $F1_{\text{last}}$ & $Acc_{\text{avg}}$ & $F1_{\text{avg}}$ & $AF$ 
 & $Acc_{\text{last}}$ & $F1_{\text{last}}$ & $Acc_{\text{avg}}$ & $F1_{\text{avg}}$ & $AF$ 
 & $Acc_{\text{last}}$ & $F1_{\text{last}}$ & $Acc_{\text{avg}}$ & $F1_{\text{avg}}$ & $AF$ 
 & $Acc_{\text{last}}$ & $F1_{\text{last}}$ & $Acc_{\text{avg}}$ & $F1_{\text{avg}}$ & $AF$ \\
\midrule
\multicolumn{21}{c}{\textit{Baseline}} \\
\midrule
JointTrain  & 88.11 & 79.95 & --    & --    & --    & 71.79 & 57.01 & --    & --    & --    & 98.83 & 98.19 & --    & --    & -- & 77.75 & 80.19 & -- & -- & -- \\
Finetune    & 35.38 & 20.97 & 53.33 & 41.45 & 43.22 & 30.19 & 9.52  & 57.97 & 31.98 & 29.27 & 43.77 & 15.85 & 66.28 & 42.34 & 37.67 & 19.66 & 12.17 & 47.56 & 41.58 & 33.69 \\
\midrule
\multicolumn{21}{c}{\textit{Regularization-based}} \\
\midrule
EWC         & 23.01 & 10.94 & 48.59 & 36.77 & 50.84 & 32.63 & 9.07  & 58.36 & 33.11 & 23.19 & 38.09 & 16.77 & 62.22 & 42.08 & 36.25 & 23.79 & 15.4 & 49.79 & 44.05 & 35.88 \\
LwF         & 36.38 & 23.56 & 59.36 & 47.01 & 29.79 & 29.47 & 15.32 & 58.23 & 38.43 & 33.44 & 24.71 & 13.37 & 43.64 & 33.55 & 20.74 & 38.98 & 26.45 & 57.57 & 49.71 & 21.00 \\
\midrule
\multicolumn{21}{c}{\textit{Replay-based (Buffer: 30 samples/class)}} \\
\midrule
iCaRL       & 73.66 & 63.08 & 82.09 & 77.37 & 23.95 & 56.36 & 40.93 & 66.63 & 52.33 & 40.77 & 74.12 & 74.64 & 88.20  & 86.72 & 32.79 & 44.80 & \underline{49.67} & 55.59 & \underline{58.86} & 45.94 \\
BiC         & 71.56 & 61.93 & 79.33 & 73.62 & 21.39 & 56.43 & 42.11 & \underline{68.61} & 52.55 & 30.55 & 73.54 & 70.44 & 87.20 & 83.90 & 37.90 & 47.80 & 45.62 & 57.23 & 55.62 & \underline{28.54} \\
WA          & 72.33 & 61.54 & 82.07 & 76.98 & 26.07 & 57.42 & 39.31 & 67.07 & 52.33 & 41.03 & 76.23 & 77.59 & 86.73 & 85.83 & 29.47 & 40.23 & 41.82 & 53.48 & 54.60 & 52.06 \\
DER         & 75.04 & 65.47 & 83.55 & 78.68 & 15.25 & 55.97 & 42.56 & 65.63 & 52.52 & 37.49 & \underline{81.60} & 79.51 & \textbf{90.62} & 87.92 & \underline{19.70} & \underline{49.85} & 49.38 & \underline{57.55} & 58.08 & 33.22 \\
FOSTER      & \underline{76.56} & \underline{68.75} & 83.91 & 79.73 & \textbf{12.59} & \underline{60.65} & \textbf{46.14} & 67.04 & 54.25 & \textbf{25.90} & 79.69 & \textbf{81.73} & 88.75 & \textbf{89.02} & \underline{19.70} & 38.18 & 40.56 & 50.13 & 52.45 & 50.99 \\
MEMO        & 73.18 & 61.93 & 81.34 & 76.14 & 21.93 & 53.07 & 37.64 & 64.84 & 51.88 & 44.09 & 77.00 & 78.78 & 87.63 & 87.05 & 26.66 & 38.80 & 41.46 & 50.02 & 51.69 & 50.48 \\
MRFA        & 72.52 & 60.21 & 82.38 & 77.40 & 26.39 & 54.71 & 38.71 & 67.78 & 53.22 & 47.34 & 75.95 & 77.52 & 87.60 & 86.47 & 29.58 & 42.79 & 48.66 & 54.04 & 57.08 & 47.64 \\
DGR         & 76.41 & \textbf{68.86} & \underline{84.67} & \underline{80.70} & 17.18 & 57.22 & 43.64 & 66.78 & \underline{54.75} & 38.38 & 80.19 & 76.92 & 88.87 & 86.44 & 23.35 & 47.11 & 48.34 & 54.52 & 56.18 & 48.67 \\
\rowcolor{gray!20}
\textbf{EndoCIL} & \textbf{78.13} & 68.45 & \textbf{85.36} & \textbf{81.38} & \underline{14.69} & \textbf{61.24} & \underline{45.94} & \textbf{71.22} & \textbf{56.06} & \underline{26.31} & \textbf{84.86} & \underline{80.85} & \underline{90.35} & \underline{87.97} & \textbf{16.24} & \textbf{57.87} & \textbf{57.53} & \textbf{65.92} & \textbf{63.06} & \textbf{19.87} \\
\midrule
\multicolumn{21}{c}{\textit{Replay-based (Buffer: 40 samples/class)}} \\
\midrule
iCaRL       & 77.70 & 65.88 & 83.44 & 78.25 & 22.57 & 57.22 & 41.17 & 68.03 & 53.64 & 42.75 & 77.47 & 76.02 & 89.47 & 88.53 & 31.49 & 47.18 & 48.32 & 55.10 & 54.73 & 39.44 \\
BiC         & 73.37 & 60.32 & 80.62 & 74.43 & 24.84 & \textbf{60.78} & \underline{45.00} & \underline{71.43} & 54.08 & \underline{28.74} & 75.53 & 71.92 & 86.61 & 84.32 & 31.12 & \underline{55.75} & \underline{57.25} & \underline{64.47} & 61.36 & \underline{18.18} \\
WA          & 74.99 & 62.95 & 83.27 & 78.33 & 24.70 & 58.93 & 39.89 & 68.39 & 53.76 & 42.92 & 76.93 & 78.08 & 89.81 & 89.03 & 27.67 & 44.80 & 48.40 & 55.57 & 57.66 & 44.15 \\
DER         & 78.27 & 67.99 & \underline{84.85} & 80.16 & 14.74 & 56.30 & 43.94 & 67.92 & 53.37 & 39.03 & 80.04 & 81.54 & 90.30 & 89.25 & 20.79 & 44.73 & 49.00 & 60.36 & \underline{61.46} & 31.57 \\
FOSTER      & \underline{78.46} & \underline{69.97} & 84.54 & \underline{80.18} & \underline{13.88} & 59.39 & 43.83 & 66.99 & 53.73 & 31.17 & 80.74 & \textbf{83.59} & \underline{90.98} & \underline{90.67} & 21.20 & 38.40 & 41.91 & 50.94 & 53.48 & 49.68 \\
MEMO        & 73.85 & 64.31 & 82.35 & 77.66 & 23.41 & 55.37 & 38.76 & 67.75 & 54.26 & 44.55 & 80.43 & 82.49 & 89.72 & 88.70 & 23.95 & 41.14 & 46.34 & 51.85 & 54.08 & 38.70 \\
MRFA        & 77.37 & 66.56 & 84.49 & 80.01 & 23.95 & 59.72 & 43.47 & 69.06 & 54.19 & 36.84 & 79.77 & 77.60 & 90.35 & 89.21 & 25.42 & 42.90 & 45.77 & 54.99 & 56.81 & 43.31 \\
DGR         & 76.70 & 68.78 & 83.14 & 78.87 & \textbf{12.22} & 55.31 & 41.76 & 67.05 & \underline{55.40} & 41.23 & \underline{81.71} & 81.15 & 89.81 & 88.03 & \underline{18.68} & 44.58 & 48.38 & 55.03 & 57.54 & 42.80 \\
\rowcolor{gray!20}
\textbf{EndoCIL} & \textbf{78.60} & \textbf{70.22} & \textbf{86.00} & \textbf{82.36} & 16.55 & \underline{60.71} & \textbf{47.00} & \textbf{71.90} & \textbf{57.23} & \textbf{24.13} & \textbf{85.49} & \underline{83.03} & \textbf{92.04} & \textbf{90.67} & \textbf{16.70} & \textbf{57.17} & \textbf{58.38} & \textbf{65.62} & \textbf{61.81} & \textbf{12.22} \\
\midrule
\multicolumn{21}{c}{\textit{Replay-based (Buffer: 50 samples/class)}} \\
\midrule
iCaRL       & 73.99 & 63.18 & 83.65 & 78.48 & 23.65 & 56.43 & 40.48 & 69.59 & 55.93 & 38.12 & 82.33 & 84.68 & 91.34 & 90.38 & 18.20 & 46.85 & 51.98 & 56.36 & 57.95 & 40.16 \\
BiC         & 77.70 & 68.71 & 82.84 & 77.07 & 16.69 & \underline{62.03} & \underline{49.12} & \textbf{73.57} & \underline{58.95} & \textbf{25.46} & 81.32 & 81.34 & 90.72 & 89.51 & 21.80 & \underline{55.05} & \underline{54.45} & \underline{63.63} & 60.27 & \underline{14.76} \\
WA          & 77.22 & 66.03 & 84.93 & 80.18 & 22.13 & 60.98 & 44.29 & 70.70 & 57.23 & 31.72 & 80.51 & 82.58 & 90.72 & 90.40 & 23.45 & 46.38 & 49.39 & 58.30 & 59.80 & 41.70 \\
DER         & \textbf{79.03} & \underline{69.35} & \underline{85.64} & \underline{81.23} & \underline{12.06} & 60.18 & 46.18 & 69.74 & 55.69 & 36.61 & 81.91 & 81.20 & 91.08 & 89.21 & 26.84 & 50.51 & 53.97 & 63.31 & \textbf{64.96} & 21.43 \\
FOSTER      & 74.66 & 65.04 & 83.91 & 79.78 & 19.00 & 59.72 & 44.10 & 67.59 & 54.23 & 33.72 & 85.84 & 85.98 & 92.44 & \underline{91.64} & \underline{16.58} & 40.59 & 43.76 & 52.44 & 55.07 & 48.40 \\
MEMO        & 76.70 & 67.16 & 83.67 & 78.79 & 19.93 & 57.61 & 46.33 & 69.01 & 56.44 & 36.71 & \underline{86.38} & \underline{86.72} & \underline{92.50} & 91.57 & 17.46 & 44.73 & 48.41 & 55.59 & 57.10 & 31.37 \\
MRFA        & 77.94 & 68.70 & 84.86 & 80.21 & 19.79 & 58.93 & 43.65 & 68.32 & 55.95 & 38.12 & 81.28 & 80.40 & 90.80 & 89.86 & 23.25 & 43.67 & 47.75 & 55.09 & 58.06 & 44.99 \\
DGR         & 73.94 & 66.05 & 84.26 & 80.12 & 20.26 & 56.03 & 43.71 & 67.34 & 56.17 & 38.12 & 83.81 & 81.41 & 90.46 & 88.63 & 18.29 & 47.80 & 51.14 & 55.76 & 58.61 & 42.33 \\
\rowcolor{gray!20}
\textbf{EndoCIL} & \underline{78.46} & \textbf{71.21} & \textbf{86.53} & \textbf{83.05} & \textbf{14.55} & \textbf{63.15} & \textbf{49.14} & \underline{72.83} & \textbf{59.35} & \underline{26.24} & \textbf{89.65} & \textbf{87.12} & \textbf{93.69} & \textbf{92.34} & \textbf{10.27} & \textbf{58.05} & \textbf{56.85} & \textbf{66.17} & \underline{62.74} & \textbf{14.47} \\
\bottomrule
\end{tabular}
}
\label{tab:results}
\end{table*}

Table~\ref{tab:results} summarizes the performance of various CIL methods on four endoscopic datasets.
JointTrain represents the upper bound performance, where all classes are available simultaneously for training with full supervision.
In contrast, Finetune denotes the lower bound, where tasks are learned sequentially without any mitigation for forgetting.
Regularization-based methods including EWC~\cite{kirkpatrick2017overcoming} and LwF~\cite{li2017learning} generally exhibit limited performance.
This demonstrates that merely constraining parameter updates is insufficient to capture complex inter-task knowledge transitions.

Replay-based methods such as iCaRL~\cite{rebuffi2017icarl}, BiC~\cite{wu2019large}, WA~\cite{zhao2020maintaining}, DER~\cite{yan2021dynamically}, FOSTER~\cite{wang2022foster}, MEMO~\cite{zhou2022model}, MRFA~\cite{zheng2024multi}, and DGR~\cite{he2024gradient} generally outperform regularization-based approaches, but their performance varies considerably across datasets and buffer settings.
BiC and FOSTER achieve relatively strong results but both methods require two-stage training during incremental sessions, increasing computational overhead.
Others, such as iCaRL and WA, show performance gaps compared to the top-performing methods.
MEMO exhibits limited generalizability, with competitive performance only observed on Kvasir-Capsule at a buffer size of 50.
DER relies on dynamic network expansion, increasing memory cost and limiting scalability.
The performance of MRFA (based on iCaRL) and DGR is also weaker than that of our method.
In contrast, EndoCIL employs a more efficient single-stage learning strategy while maintaining superior performance across diverse scenarios.
It ranks first or second across most metrics on all four datasets and buffer settings.
For instance, with a buffer size of 30, EndoCIL achieves the highest $Acc_{\text{last}}$ on four datasets, outperforming the second-best methods by 1.57, 0.59, 3.26 and 8.02.
It also achieves the highest $Acc_{\text{avg}}$ with a buffer size of 40, surpassing the second-best methods by 1.15, 0.47, 1.06 and 1.15.
On the Capsule Vision dataset, EndoCIL achieves top results across all buffer sizes, ranking first in all but one metric.
Overall, EndoCIL demonstrates strong competitiveness across all datasets and buffer sizes, highlighting its effectiveness for endoscopic class-incremental learning.

\begin{figure*}[h]
    \centering
    \includegraphics[width=\textwidth]{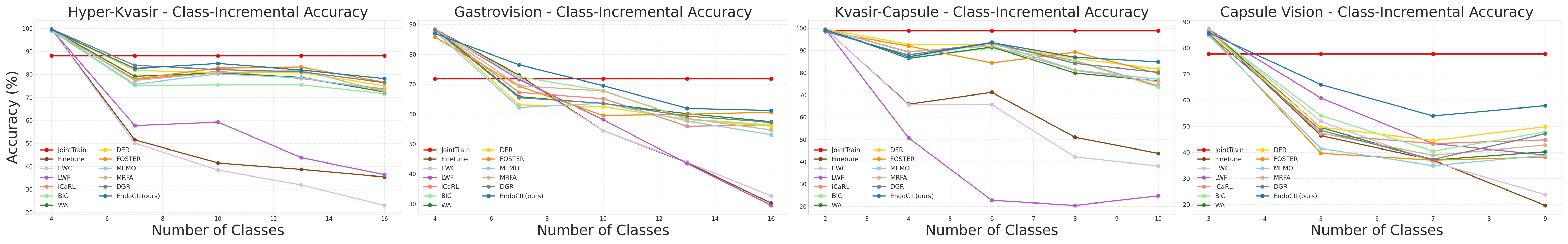}
    \caption{Performances of different methods across learning sessions on four endoscopic datasets.}
    \label{img:Performances}
    \vspace{-1.0em}
\end{figure*}

As illustrated in Fig.~\ref{img:Performances}, we further present the incremental accuracy trends of various methods under a buffer size of 30.
EndoCIL consistently achieves superior performance across all incremental tasks on Gastrovision and Capsule Vision, and is only slightly outperformed by DGR on the first and FOSTER on the third task of Hyper-Kvasir, while leading in the others.
On Kvasir-Capsule, although EndoCIL shows lower accuracy than some baselines on tasks 1 and 3, it clearly outperforms them on the final task.
These results across four datasets and multiple tasks highlight the stability and generalization ability of EndoCIL.

\subsection{Ablation Study}
\label{sec:ablation}
To better quantify the individual contribution of each proposed component, we conducted an ablation study on the Hyper-Kvasir dataset. 
We adopt iCaRL as the baseline, which employs mean-based replay and knowledge distillation to alleviate forgetting.

\renewcommand{\thetable}{3}
\begin{table*}[htbp]

\centering
\caption{Ablation study results on the Hyper-Kvasir dataset.}
\begin{tabular}{ccc|ccccc}
\toprule
\textbf{MDBR} & \textbf{PRCBL} & \textbf{CFG} & 
$Acc_{\text{last}}$ & $F1_{\text{last}}$ & $Acc_{\text{avg}}$ & $F1_{\text{avg}}$ & $AF$ \\
\midrule
\xmark        & \xmark               & \xmark        & 73.66 & 63.08 & 82.09 & 77.37 & 23.95 \\
\checkmark    & \xmark               & \xmark        & 75.27 & 63.11 & 83.11 & 77.90 & 24.67 \\
\xmark        & \checkmark           & \xmark        & 75.42 & 68.16 & 83.71 & 79.62 & 13.50 \\
\xmark        & \xmark               & \checkmark    & 69.95 & 60.05 & 80.69 & 75.44 & 27.68 \\
\checkmark    & \checkmark           & \xmark        & 76.99 & \textbf{69.38} & 84.29 & 80.48 & 15.12 \\
\checkmark    & \xmark               & \checkmark    & 69.47 & 60.45 & 80.74 & 76.03 & 30.95 \\
\xmark        & \checkmark           & \checkmark    & 76.65 & 67.72 & 84.14 & 79.81 & \textbf{13.40} \\
\checkmark    & \checkmark           & \checkmark    & \textbf{78.13} & 68.45 & \textbf{85.36} & \textbf{81.38} & 14.69 \\
\bottomrule
\end{tabular}
\label{table:ablation}
\end{table*}

As shown in Table~\ref{table:ablation}, although combining multiple components occasionally leads to fluctuations in individual metrics, the fully optimized model achieves the highest scores across key indicators. 
Compared to the baseline, it improves $Acc_{\text{last}}$, $F1_{\text{last}}$, $Acc_{\text{avg}}$, and $F1_{\text{avg}}$ by 4.47, 5.37, 3.27, and 4.01, while also reducing $AF$ by 9.26.
Introducing any single component or combining two of them generally yields 1-3 point improvements over the baseline on both $Acc_{\text{last}}$ and $Acc_{\text{avg}}$.
Notably, combining CFG with the original cross-entropy (CE) loss leads to performance degradation, whereas integrating CFG with PRCBL brings significant gains.
This is because directly applying gradient calibration to the FC layer under imbalanced loss leads to underfitting of new classes, which we analyze in section~\ref{sec:gradient}.
Overall, these results highlight the effectiveness of each component in balancing stability and plasticity.

\subsection{Sample Selection Analysis}

\begin{figure*}[h]
    \centering
    \includegraphics[width=\textwidth]{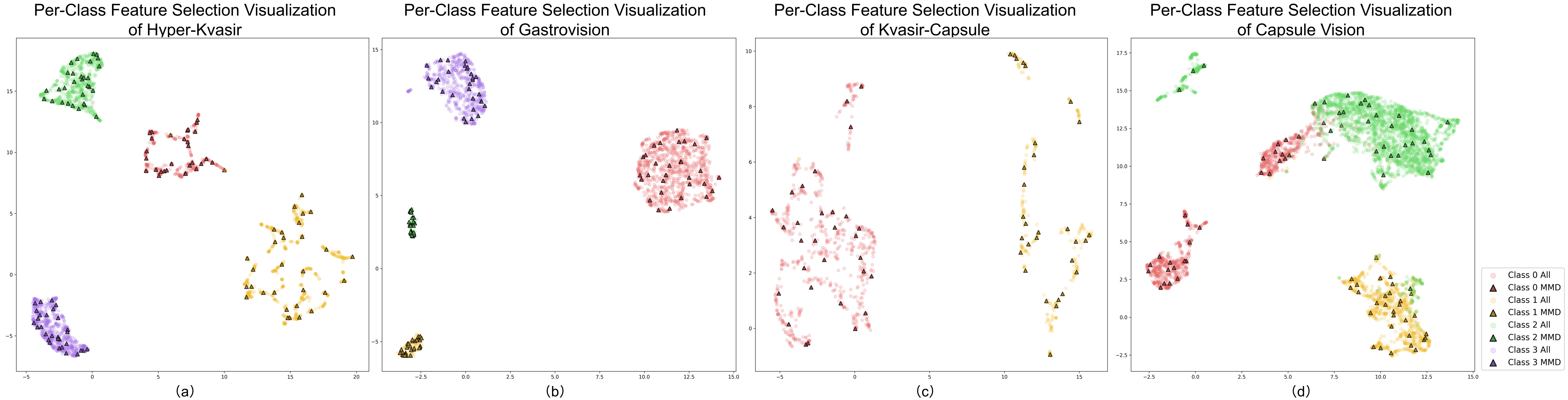}
    \caption{UMAP-based visualization of feature distributions for MMD-based selection strategy across four endoscopic datasets. Light circles ($\circ$) denote all training samples and triangles ($\blacktriangle$) indicate selections by MMD-based method.}
    \label{img:visualization}
\end{figure*}

To intuitively show the MMD-based selection method, we visualize the distribution of selected features using UMAP~\cite{mcinnes2018umap}.
To eliminate the influence of feature extractor changes caused by different replay strategies in subsequent tasks, we fix the feature extractor using the model obtained at task 0 of iCaRL.
Based on this fixed extractor, we collect training features of task 0 classes from four endoscopic datasets.
As shown in Fig.~\ref{img:visualization}, the MMD-based selection strategy offers broad coverage of the overall feature space, preserving diverse samples across the distribution.

\begin{figure*}[h]
    \centering
    \includegraphics[width=\textwidth]{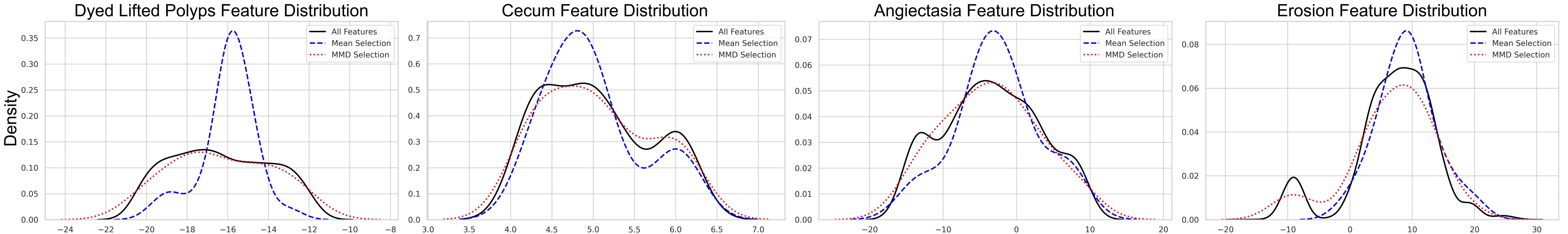}
    \caption{One-dimensional KDE distributions of original and replay samples across four classes in different endoscopic datasets.}
    \label{img:distribution}
\end{figure*}

To further analyze the differences between the Mean and MMD selection strategies and validate the effectiveness of our method in capturing the original feature distribution, we reduce their features to one dimension using UMAP and estimate their distributions using Kernel Density Estimation (KDE)~\cite{davis2011remarks, parzen1962estimation}.
As shown in Fig.~\ref{img:distribution}, samples selected by the MMD-based strategy exhibit a distribution that more closely aligns with that of the original features, while samples selected by the Mean-based method are heavily concentrated near the peak density of the original distribution.
This indicates that the Mean-based method tends to overlook feature diversity, particularly in low-density boundary regions.
We argue that such a strategy, which focuses solely on approximating the class mean, inherently limits the diversity and richness of stored exemplars, thereby exacerbating the issue of catastrophic forgetting.

\subsection{Loss Function Analysis}

\begin{figure}[h]
    \centering
    \includegraphics[width=0.7\textwidth]{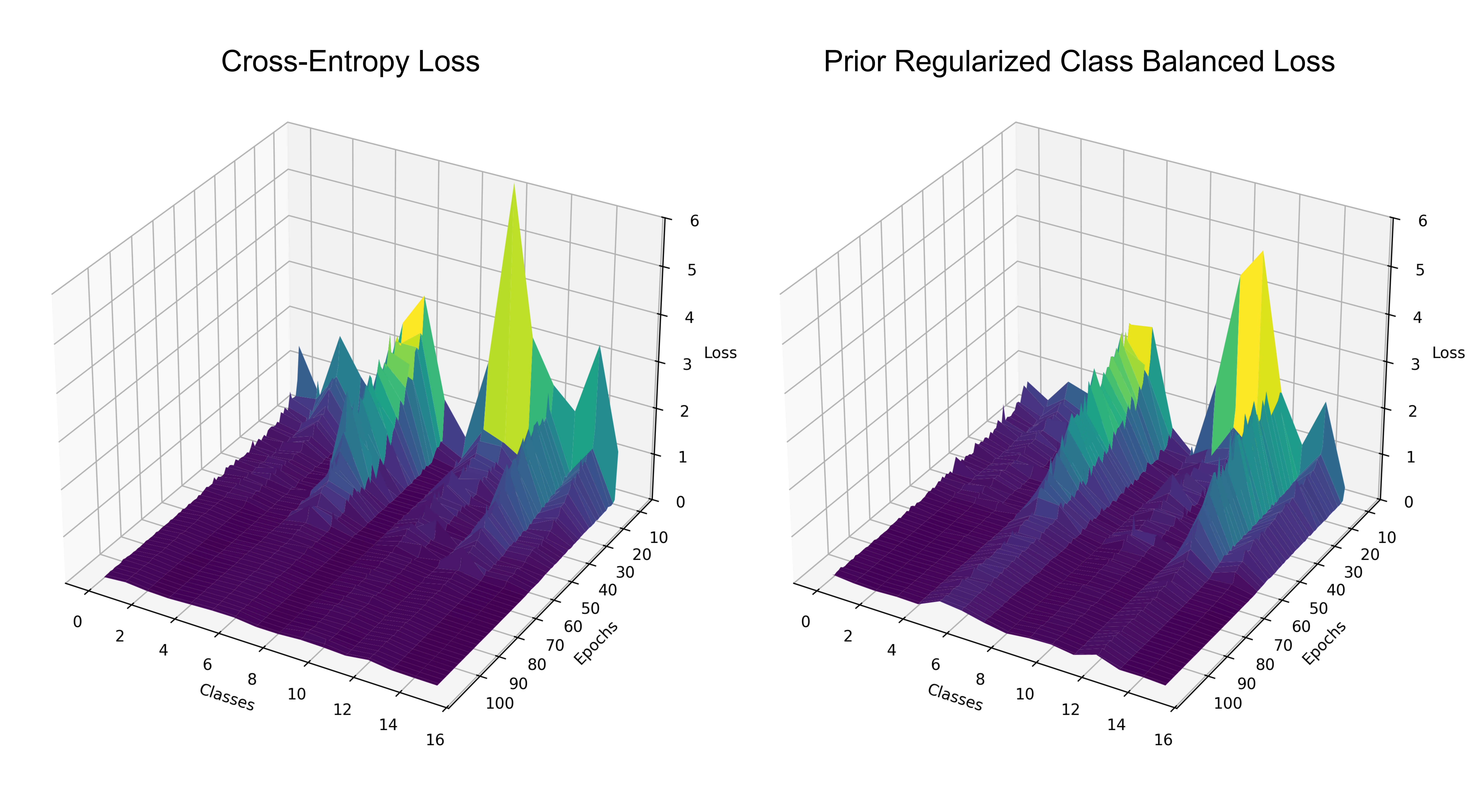}
    \caption{Visualization comparison of Cross-Entropy Loss and Prior Regularized Class Balanced Loss distributions on Task 4 of the Hyper-Kvasir dataset. The x-axis represents the class, and the y-axis represents the epoch.}
    \label{img:loss_clf}
\end{figure}

To directly demonstrate that PRCBL can effectively alleviate dual-class imbalance, we use iCaRL as the baseline to visualize the distributions of CE Loss and PRCBL across classes on Task 4 of the Hyper-Kvasir dataset.
As shown in Fig.~\ref{img:loss_clf}, during the early stages of training, the overall distribution of PRCBL across different classes is significantly more balanced than that of CE Loss.
Specifically, the loss values for classes 0, 2, 5, 9, and 13 are notably lower.
This indicates that the model guided by PRCBL is better at adapting to new tasks, with less impact on previous tasks compared to the traditional CE Loss.

\begin{table*}[ht]
\centering
\caption{Comparison of accuracy for two kinds of loss across different classes on Task 4 of the Hyper-Kvasir dataset.}
\setlength{\tabcolsep}{0.5mm}
\resizebox{\textwidth}{!}{
\begin{tabular}{cccccccccccccccccc}
\hline
\textbf{Method} & \textbf{Class 0} & \textbf{Class 1} & \textbf{Class 2} & \textbf{Class 3} & \textbf{Class 4} & \textbf{Class 5} & \textbf{Class 6} & \textbf{Class 7} & \textbf{Class 8} & \textbf{Class 9} & \textbf{Class 10} & \textbf{Class 11} & \textbf{Class 12} & \textbf{Class 13} & \textbf{Class 14} & \textbf{Class 15} & \textbf{Average Acc} \\
\hline
CE Loss & 83.85 & 75.22 & 80.69 & \textbf{59.20} & 62.63 & 1.23 & 36.54 & \textbf{100.00} & 87.86 & \textbf{89.00} & 74.68 & 88.89 & 2.44 & \textbf{62.92} & \textbf{66.67} & \textbf{98.93} & 66.92 \\
PRCBL & \textbf{86.15} & \textbf{83.04} & \textbf{94.55} & 44.28 & \textbf{82.83} & \textbf{23.46} & \textbf{55.77} & 96.30 & \textbf{95.63} & 88.00 & \textbf{87.34} & \textbf{91.50} & \textbf{24.39} & 52.81 & 40.74 & 97.33 & \textbf{71.51} \\
\hline
\end{tabular}
}
\label{tab:acc_comparison}
\end{table*}

We also recorded the performance of two different loss functions on Task 4 of the Hyper-Kvasir dataset across each class.
As shown in Table~\ref{tab:acc_comparison}, the average accuracy of the PRCBL-guided model is superior to that of the CE loss, with a generally better performance across all classes.
Particularly, there is a noticeable performance gap in class 2, 4, 5, 6, and 12.
It is worth noting that the model guided by CE loss performs poorly on class 5 and 12 ($ACC < 3$), making it nearly impossible to effectively classify images from these classes.
In contrast, the PRCBL-guided model retains some level of recognition capability for these classes, significantly outperforming the CE loss-based model.
However, in terms of recognizing the new classes in Task 4 (Classes 13–15), the PRCBL model underperforms compared to the CE loss model, indicating that PRCBL still has limitations in balancing the performance between new and old tasks.

\subsection{Fully Connected Layer Gradients Analysis}
\label{sec:gradient}
As mentioned in~\ref{sec:ablation}, directly combining CFG with CE and distillation loss results in a performance drop.
To investigate the underlying cause, we use iCaRL as the baseline and visualize the changes in the L2 norm of the gradient of the FC layer weights at Task 4, comparing the cases where CFG is applied directly and where CFG is combined with PRCBL. 

\begin{figure}[h]
    \centering
    \includegraphics[width=0.9\textwidth]{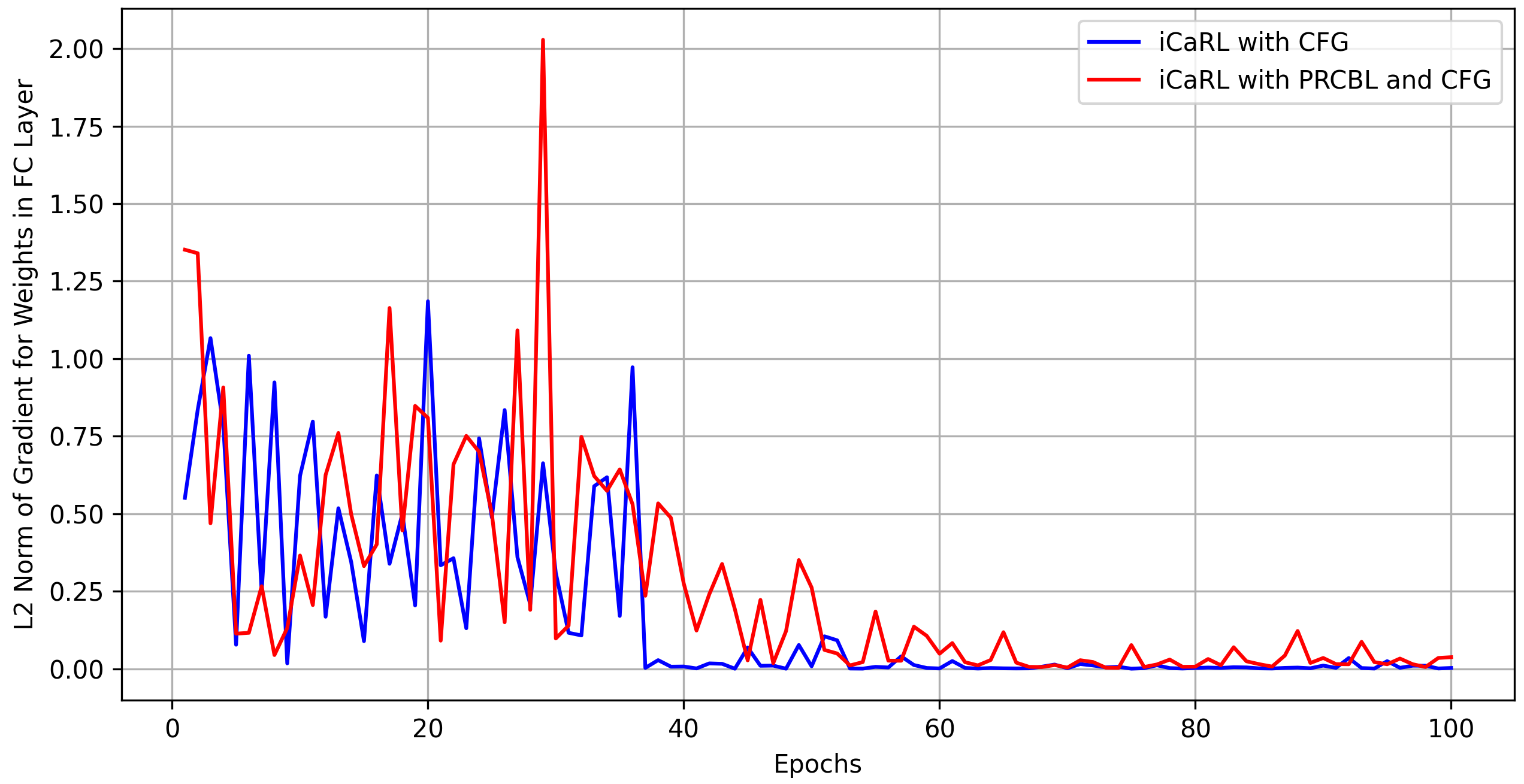}
    \caption{Comparison of L2 norm of gradient for weights in FC layer across epochs on Task 4 of the Hyper-Kvasir dataset.}
    \label{img:gradient}
\end{figure}

As shown in Fig.~\ref{img:gradient}, when CFG is applied directly, the gradient of the FC layer drops to near zero around 40 epochs and remains close to zero for the subsequent epochs.
In contrast, when CFG is combined with PRCBL, this issue is alleviated. 
Specifically, the gradient of the FC layer decreases to a low value around 60 epochs but continues to experience small fluctuations near zero thereafter. 
This phenomenon can be explained by the design of CFG, which aims to attenuate the gradients of new classes in order to mitigate the classifier's bias towards them.
However, directly applying CFG may excessively suppress the gradient variations of new classes, causing the FC layer weights to stop updating prematurely and leading to underfitting of these classes. 
By integrating CFG with PRCBL, the gradients of new classes are moderately attenuated, which allows the FC layer to continue updating and ensures that CFG effectively prevents the classification layer from disproportionately prioritizing new classes.

\section{Conclusion}
In this paper, we proposed EndoCIL, a specialized class-incremental learning framework for endoscopic image analysis. 
To tackle catastrophic forgetting caused by domain discrepancies and class imbalance in real-world scenarios, we designed three key components: MDBR to ensure the diversity and representativeness of stored exemplars, PRCBL to mitigate dual-phase class imbalance, and CFG to reduce classifier bias toward new classes.
Experiments on four public endoscopic datasets demonstrate its superior performance over existing state-of-the-art CIL methods.
By combining continual learning principles with the unique characteristics of endoscopic imagery, EndoCIL offers a robust foundation for scalable, lifelong diagnostic systems.

\section*{Acknowledgements}
This work was partly supported by the Anhui Provincial Natural Science Foundation (Grant No. 2408085MF162) and the Beijing Natural Science Foundation (Grant No. 7242270).

\bibliographystyle{unsrt}  
\bibliography{EndoCIL}

\end{document}